\documentclass{article}

\usepackage[dblblindworkshop,final]{neurips_2025}

\workshoptitle{MATH-AI}

\usepackage[utf8]{inputenc} 
\usepackage[T1]{fontenc}    
\usepackage{hyperref}       
\usepackage{url}            
\usepackage{booktabs}       
\usepackage{amsfonts}       
\usepackage{nicefrac}       
\usepackage{microtype}      
\usepackage{xcolor}         

\usepackage{kotex}

\usepackage{amsmath}
\usepackage{amssymb}
\usepackage{mathtools}
\usepackage{amsthm}
\usepackage{thmtools,thm-restate}
\theoremstyle{definition}
\newtheorem{definition}{Definition}
\newtheorem{theorem}{Theorem}
\usepackage[pdftex]{graphicx}
\usepackage[capitalise]{cleveref}

\usepackage{booktabs}
\usepackage{siunitx}
\usepackage{multirow}

\title{Numbers Already Carry Their Own Embeddings}

\author{%
  Suhyun Bae \And
  Donghun Lee\thanks{Corresponding author.} \\
  \AND
  Department of Mathematics \\
  Korea University \\
  \texttt{\{baeshstar, holy\}@korea.ac.kr}
}

\begin{document}

\maketitle

\begin{abstract}
We introduce \textbf{Adelic operation-preserved embeddings (AOE)}, a training-free representation that captures both a number’s real value and its modular ($p$-adic) signatures. 
This construction preserves additive and multiplicative structure by design, turning numerical input into embeddings that “speak in the language of mathematics.” 
Unlike prior approaches that rely on task-specific retraining, AOE is plug-and-play and drops seamlessly into existing architectures. 
On algebraic combinatorics benchmarks, it delivers consistent gains—including the first-ever perfect accuracy on the Weaving Pattern task—while suggesting a principled path forward for overcoming the long-standing “number problem” in AI.
\end{abstract}


\section{Introduction}
\label{intro}
LLMs have recently crossed remarkable milestones in mathematical reasoning—for instance, Gemini 2.5 Pro attained gold medal–level performance at the International Mathematical Olympiad (IMO) 2025~\cite{huang2025gemini25procapable}.
Yet a paradox has emerged: on certain core tasks, simpler baseline models like vanilla transformers or even MLPs can outperform state-of-the-art systems such as Claude 3.5 Sonnet and GPT-4o~\cite{chau2025machine}.
This surprising reversal highlights that scale alone has not solved one of the oldest bottlenecks in AI: \textbf{numerical representation}.

Most LLMs still treat numbers as strings of characters. 
From Byte Pair Encoding (BPE)~\cite{BPE} to modern tokenization schemes, digits are fragmented into arbitrary subword units, stripping away the additive and multiplicative relationships that define numbers themselves~\cite{wallace-etal-2019-nlp, thawani2021representingnumbersnlpsurvey}. 
The result is brittle reasoning, inconsistent generalization, and systematic errors in tasks that require arithmetic precision.
Prior efforts have explored continuous embeddings, digit-wise positional encodings, or symbolic pretraining~\cite{golkar2024xvalcontinuousnumericaltokenization, levy2025languagemodelsencodenumbers}.
While valuable, these methods rely on the hope that models \textit{implicitly} rediscover mathematical rules during training, rather than embedding them directly.

We argue for a different path: numbers already carry their own embeddings. 
Drawing on algebraic number theory, we introduce Adelic operation-preserved embeddings (AOE), a training-free representation that encodes each number by combining its real value with modular ($p$-adic) expansions. 
This construction preserves both additive and multiplicative structures by design, providing a principled alternative to ad hoc tokenization.

\paragraph{Contributions} Our contributions are as follows:
\begin{enumerate}
    \item We propose AOE, a training-free, algebraically grounded representation that directly respects numerical structure.
    \item We frame AOE as a step toward resolving the long-standing “number problem” in AI, offering a plug-and-play remedy applicable across architectures.
    \item On algebraic combinatorics benchmarks, AOE-equipped Transformers outperform baselines on every task, notably achieving the first-ever perfect accuracy on the Weaving Pattern benchmark.
\end{enumerate}

\section{Method}
\label{method}

Our goal is to design a numerical representation that preserves the algebraic structure already inherent in numbers. 
Rather than fragmenting numbers into subword tokens, we construct embeddings directly in an \emph{Adele space}, which combines a real component with multiple $p$-adic components. 
Each $p$-adic component lies in $\mathbb{Z}_p$, a ring with well-defined addition and multiplication inherited from $\mathbb{Z}$. 
As a result, this yields a product structure where arithmetic is preserved coordinate-wise: for rationals $q_1, q_2$,
$$\mathbf{A}(q_1+q_2) = \mathbf{A}(q_1) \oplus \mathbf{A}(q_2), \quad \mathbf{A}(q_1\cdot q_2) = \mathbf{A}(q_1) \otimes \mathbf{A}(q_2)$$
where $\oplus,\,\otimes$ denote coordinate-wise operations on concatenated real and $p$-adic coordinates implemented via $N$-digit truncations.

\paragraph{Adele Space.}
The Adele Ring, a classical construction from algebraic number theory~\cite{Lang1994}, unifies the real line with all $p$-adic completions of the rationals. 
Intuitively, each number carries a distinct ``signature'' across different primes, which AOE captures alongside its real value. 
For example, \nicefrac{7}{5} is represented not only as $1.4$ in $\mathbb{R}$ but also through $p$-adic expansions, such as $\dots11011$ in the 2-adics or $\dots01212$ in the 3-adics. 
These complementary perspectives allow the embedding to encode both magnitude and modular structure.

\paragraph{Embedding Construction.}
To build the embedding of a rational number $q$, we concatenate:  
(i) a real-valued component, and  
(ii) a finite set of $p$-adic expansions truncated to $N$ digits of precision.  
The resulting tensor has shape $(n_p+1) \times N$, where $n_p$ is the number of chosen primes. 
Implementation details—such as digit lifting with Hensel’s Lemma and exact $p$-adic expansions—are provided in the \cref{sec:def} and \cref{Hensel}

\begin{figure}[h]
    \centering
    \includegraphics[height=0.3\textheight]{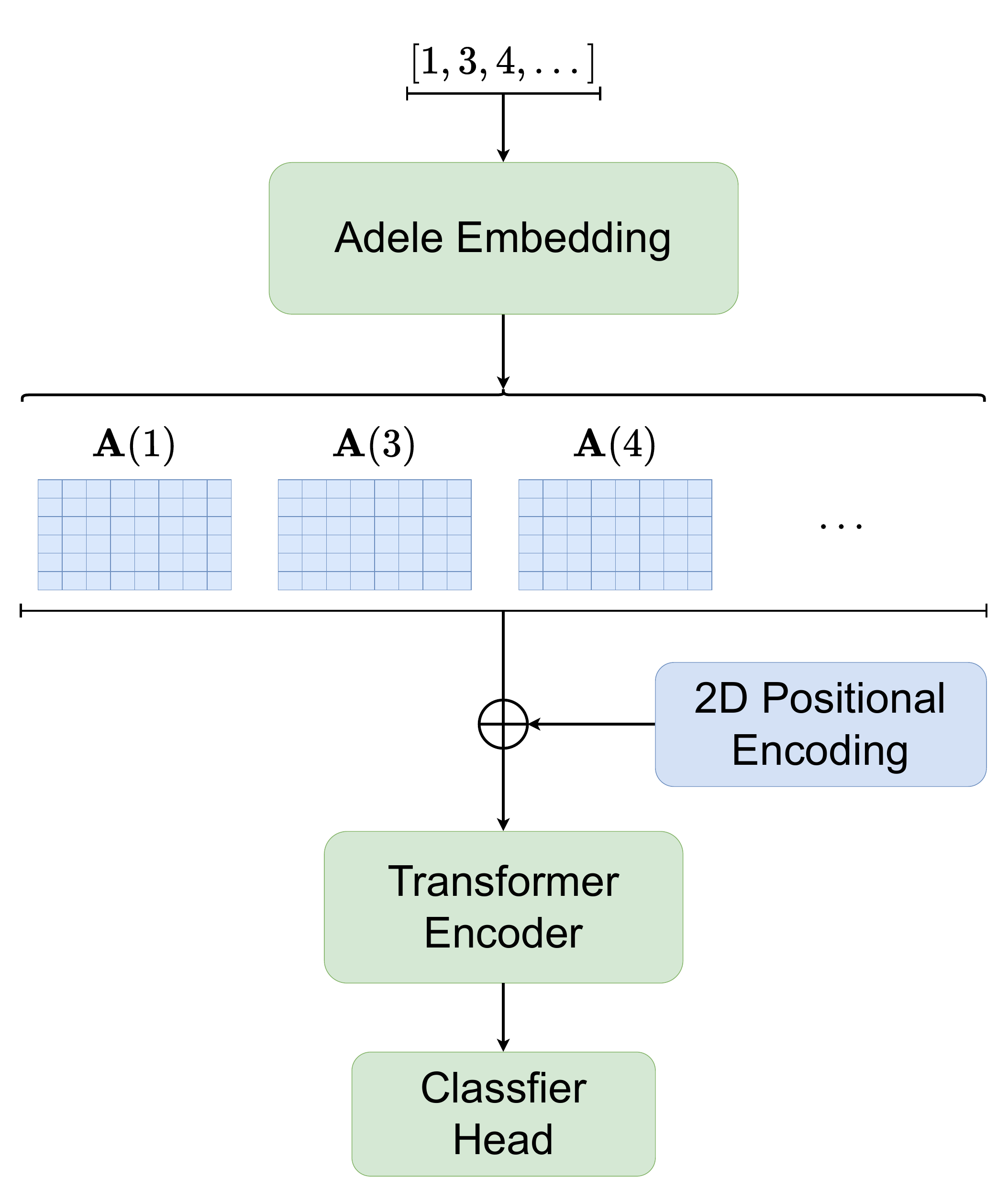}
    \caption{Diagram summarizing AOE-equipped Transformer model workflow architecture for experiments.}
    \label{fig:architecture}
\end{figure}

\paragraph{Plug-and-play Integration.}
AOE is training-free and architecture-agnostic. 
It can directly replace standard embedding layers in Transformers or other neural models without task-specific tuning. 
In practice, this means the input embedding lookup is replaced with pre-computed Adelic representation tensors, which are flattened into $d_{\text{model}}$-dimensional vectors. 
We also add a lightweight 2D positional encoding to incorporate both the sequence position and the internal structure of each Adelic tensor. 
The resulting pipeline integrates seamlessly with standard Transformer encoders, as illustrated in \cref{fig:architecture}.

\section{Experiments and Results}

\subsection{Models}
To isolate the effect of input representation, we implement two Transformer encoder models that are identical in architecture but differ in their embedding layers. 
Both models use a 6-layer Transformer encoder with 8 attention heads, hidden dimension $d_{\text{model}}=128$, and a dropout rate of 0.1. 
The final hidden state corresponding to the [CLS] token is passed through a linear layer for classification. 

The \textbf{Baseline} model employs a standard \texttt{nn.Embedding} lookup for integer tokens, followed by sinusoidal positional encodings. 
Because \texttt{nn.Embedding} is trainable, this baseline introduces additional learnable parameters. 
By contrast, the \textbf{AOE-equipped} model replaces the embedding lookup with pre-computed Adelic representation tensors, which are training-free and flattened into $d_{\text{model}}$-dimensional vectors. 
We also add a 2D positional encoding that jointly encodes the sequence index, the prime index, and the digit index of each $p$-adic expansion, thereby preserving structural information. 

Crucially, this design ensures that any performance differences arise solely from the representation, not from model capacity: the baseline in fact has more learnable parameters, making AOE's improvements all the more significant.

\subsection{Dataset}

We evaluate our approach on the \textbf{Algebraic Combinatorics Dataset (ACD)}~\cite{chau2025machine}, a benchmark designed to support the conjecturing phase of mathematical research. 
Unlike benchmarks such as \verb|GSM8K|~\cite{cobbe2021trainingverifierssolvemath} or other arithmetic word-problem datasets—which primarily focus on middle or elementary school mathematics—ACD incorporates tasks linked to open problems in algebraic combinatorics. 
It is therefore a uniquely demanding testbed: it not only exhibits scale and imbalance (up to $10^7$ instances with heavy-tailed distributions) but also requires models to reason over algebraic structures well beyond rote calculation. 

From the nine tasks available in the repository, we focus on six classification problems spanning permutations, tableaux, lattice paths, quivers, and polynomial invariants. 
This selection provides both breadth of combinatorial objects and diversity of input types (strings, matrices, graphs). 
Detailed task descriptions and mathematical context are deferred to Appendix~\ref{app:dataset}.

\subsection{Results}
\label{results}

Both models are trained from scratch with the Adam optimizer and Cross-Entropy loss. 
To address class imbalance in the ACD tasks, we use a \verb+WeightedRandomSampler+ to balance training batches. 
Additional hyperparameters, learning rate schedules, and per-task settings are reported in \cref{app:training}.

Across all six tasks, AOE consistently outperforms the baseline. 
This supports our central claim that \emph{explicitly encoding algebraic structure enables stronger generalization than text-based embeddings}. 

The improvements range from modest but consistent gains (e.g., lattice path classification) to dramatic jumps (e.g., Schubert polynomials and quiver mutation classes). 
Most strikingly, our model achieves the first-ever perfect accuracy on the Weaving Patterns benchmark, a task that has resisted even large-scale language models. 

These findings suggest that algebraically grounded representations can close long-standing gaps in numerical reasoning. 
Rather than relying on scale or implicit learning, embedding numbers in a mathematically faithful space offers a principled path to more reliable symbolic reasoning in neural models.

\cref{tab:main_acc} and \cref{tab:main_loss} provide detailed accuracy and loss values.

\begin{table}[ht]
\centering
\caption{Final test accuracy (\%) comparison on the Algebraic Combinatorics benchmark. Higher is better.}
\label{tab:main_acc}
\begin{tabular}{llrr}
\toprule
\multicolumn{2}{c}{\textbf{Dataset Task}} & \multicolumn{2}{c}{\textbf{Models}} \\
\cmidrule(lr){1-2} \cmidrule(lr){3-4}
\textbf{Task} & \textbf{$n$} & \textbf{Baseline} & \textbf{Ours} \\
\midrule

\multirow[t]{3}{*}{Lattice paths} 
    & 10 & 66.19 & 70.10 \\
    & 11 & 66.30 & 66.30 \\
    & 12 & 66.50 & 78.44 \\
\midrule
\multirow[t]{2}{*}{Weaving patterns} 
    & 6  & 53.02 & 100.00 \\
    & 7  & 51.53 & 100.00 \\
\midrule
Quiver mutation classes &   & 45.13 & 91.11 \rule{0pt}{2.5ex}\\
\midrule
Grassmannian cluster algebras &   & 91.27 & 97.39 \rule{0pt}{2.5ex}\\
\midrule
\multirow[t]{3}{*}{Schubert polynomials} 
    & 4  & 50.59 & 87.89 \\
    & 5  & 49.83 & 96.87 \\
    & 6  & 50.00  & 99.96 \\
\midrule
\multirow[t]{3}{*}{mHeight} 
    & 8  & 91.42 & 96.94 \\
    & 9  & 93.20 & 99.17 \\
    & 10 & 94.15 & 99.66 \\

\bottomrule
\end{tabular}
\end{table}

\section{Discussion}
Our current framework is restricted to rational numbers, since the Adele Ring is defined over $\mathbb{Q}$. 
Extending the construction to irrationals, complex values, or transcendental constants requires new theoretical tools. 

Another limitation concerns the manual choice of primes and digit precision. 
In this study, we fix these hyperparameters heuristically. 
More principled or adaptive strategies may improve both efficiency and expressivity. 

Finally, while AOE is training-free, its use introduces computational overhead: pre-computations and 2D positional encodings make training roughly $4$–$5$ times slower than standard embeddings. 

We view these challenges not as barriers but as opportunities for future work in broadening AOE toward more general and scalable settings.

\section{Conclusion}
This work begins from a simple observation: numbers already contain the structure they need to be represented. 
By making this structure explicit through AOE, we show that neural models achieve levels of reliability on symbolic mathematical tasks that elude even the largest LLMs. 

The broader message is conceptual. 
If numbers are more than tokens, then the future of AI reasoning hinges on embedding \emph{mathematical structures of objects}, not merely the language that describes them. 
AOE exemplifies how ideas from algebraic number theory operate as plug-and-play components in neural architectures, suggesting a path toward uniting symbolic rigor with neural scalability.

\begin{ack}
This work was supported by the National Research Foundation of Korea (NRF) grant funded by the Korea government (MSIT) (RS-2025-24873052).
\end{ack}

\medskip
\newpage


{
\small
\bibliographystyle{unsrt}
\bibliography{refs.bib}

@inproceedings{
    chau2025machine,
    title={Machine Learning meets Algebraic Combinatorics: A Suite of Datasets Capturing Research-level Conjecturing Ability in Pure Mathematics},
    author={Herman Chau and Helen Jenne and Davis Brown and Jesse He and Mark Raugas and Sara C. Billey and Henry Kvinge},
    booktitle={Forty-second International Conference on Machine Learning},
    year={2025},
    url={https://openreview.net/forum?id=tlniJJFUW2}
}

@Inbook{
    Lang1994,
    author="Lang, Serge",
    title="Ideles and Adeles",
    bookTitle="Algebraic Number Theory",
    year="1994",
    publisher="Springer New York",
    address="New York, NY",
    pages="137--154",
    isbn="978-1-4612-0853-2",
    doi="10.1007/978-1-4612-0853-2_7",
    url="https://doi.org/10.1007/978-1-4612-0853-2_7"
}

@article{Ostrowski1916,
  author    = {Ostrowski, Alexander},
  title     = {Über einige Lösungen der Funktionalgleichung φ(x)·φ(y)=φ(xy)},
  journal   = {Acta Mathematica},
  volume    = {41},
  number    = {1},
  pages     = {271–284},
  year      = {1916},
  doi       = {10.1007/BF02422947},
  issn      = {0001-5962},
  edition   = {2nd}
}

@book{Hensel1908,
  author    = {Hensel, Kurt},
  title     = {Theorie der algebraischen Zahlen},
  publisher = {B. G. Teubner},
  year      = {1908},
  address   = {Leipzig},
  url       = {https://archive.org/details/in.ernet.dli.2015.493154}
}

@misc{huang2025gemini25procapable,
      title={Gemini 2.5 Pro Capable of Winning Gold at IMO 2025}, 
      author={Yichen Huang and Lin F. Yang},
      year={2025},
      eprint={2507.15855},
      archivePrefix={arXiv},
      primaryClass={cs.AI},
      url={https://arxiv.org/abs/2507.15855}, 
}

@article{BPE,
author = {Gage, Philip},
title = {A new algorithm for data compression},
year = {1994},
issue_date = {Feb. 1994},
publisher = {R \& D Publications, Inc.},
address = {USA},
volume = {12},
number = {2},
issn = {0898-9788},
journal = {C Users J.},
month = feb,
pages = {23–38},
numpages = {16}
}

@inproceedings{wallace-etal-2019-nlp,
    title = "Do {NLP} Models Know Numbers? Probing Numeracy in Embeddings",
    author = "Wallace, Eric  and
      Wang, Yizhong  and
      Li, Sujian  and
      Singh, Sameer  and
      Gardner, Matt",
    editor = "Inui, Kentaro  and
      Jiang, Jing  and
      Ng, Vincent  and
      Wan, Xiaojun",
    booktitle = "Proceedings of the 2019 Conference on Empirical Methods in Natural Language Processing and the 9th International Joint Conference on Natural Language Processing (EMNLP-IJCNLP)",
    month = nov,
    year = "2019",
    address = "Hong Kong, China",
    publisher = "Association for Computational Linguistics",
    url = "https://aclanthology.org/D19-1534/",
    doi = "10.18653/v1/D19-1534",
    pages = "5307--5315"
}

@misc{thawani2021representingnumbersnlpsurvey,
      title={Representing Numbers in NLP: a Survey and a Vision}, 
      author={Avijit Thawani and Jay Pujara and Pedro A. Szekely and Filip Ilievski},
      year={2021},
      eprint={2103.13136},
      archivePrefix={arXiv},
      primaryClass={cs.CL},
      url={https://arxiv.org/abs/2103.13136}, 
}

@misc{golkar2024xvalcontinuousnumericaltokenization,
      title={xVal: A Continuous Numerical Tokenization for Scientific Language Models}, 
      author={Siavash Golkar and Mariel Pettee and Michael Eickenberg and Alberto Bietti and Miles Cranmer and Geraud Krawezik and Francois Lanusse and Michael McCabe and Ruben Ohana and Liam Parker and Bruno Régaldo-Saint Blancard and Tiberiu Tesileanu and Kyunghyun Cho and Shirley Ho},
      year={2024},
      eprint={2310.02989},
      archivePrefix={arXiv},
      primaryClass={stat.ML},
      url={https://arxiv.org/abs/2310.02989}, 
}

@misc{levy2025languagemodelsencodenumbers,
      title={Language Models Encode Numbers Using Digit Representations in Base 10}, 
      author={Amit Arnold Levy and Mor Geva},
      year={2025},
      eprint={2410.11781},
      archivePrefix={arXiv},
      primaryClass={cs.LG},
      url={https://arxiv.org/abs/2410.11781}, 
}

@misc{cobbe2021trainingverifierssolvemath,
      title={Training Verifiers to Solve Math Word Problems}, 
      author={Karl Cobbe and Vineet Kosaraju and Mohammad Bavarian and Mark Chen and Heewoo Jun and Lukasz Kaiser and Matthias Plappert and Jerry Tworek and Jacob Hilton and Reiichiro Nakano and Christopher Hesse and John Schulman},
      year={2021},
      eprint={2110.14168},
      archivePrefix={arXiv},
      primaryClass={cs.LG},
      url={https://arxiv.org/abs/2110.14168}, 
}
}

\newpage
\appendix
\appendix
\section{Formalization of Adelic Operation-preserved Embedding}
\label{sec:def}
\subsection{Adele Ring}
\begin{definition}[Adele Ring]
    Let $K$ be a global field, and $\mathbb{A}_K$ be an "\textbf{Adele Ring}" if 
    $$ \mathbb{A}_K = \prod_{v} (K_v, \mathcal{O}_v) $$
    where $v$ represents all possible valuations on $K$.
\end{definition}

For the field of rationals $K = \mathbb{Q}$, \textbf{Ostrowski's theorem}~\cite{Ostrowski1916} implies that all non-trivial valuations correspond either to the reals $\mathbb{R}$ or to $p$-adic integers $\mathbb{Z}_p$ for some prime $p$. Hence:

$$ \mathbb{A}_{\mathbb{Q}} = \mathbb{R} \times \prod_{p} (\mathbb{Q}_p, \mathbb{Z}_p).$$

An element $a \in \mathbb{A}_{\mathbb{Q}}$ can be written as an infinite tuple:

$$ a = (a_{\infty}, a_2, a_3, a_5, a_7, \dots).$$

\subsection{Example Representation}
For instance, the rational number $\tfrac{7}{5}$ can be represented as:

$$ \mathbf{A}_{\mathbb{Q}}\left( \frac{7}{5} \right) = (1.4, ~\dots 11011_2, ~\dots 01212_3, 1.2_5, ~\dots 54130_7, ~\dots ) $$

where each $p$-adic component is approximated with an $N$-digit precision.

\subsection{Embedding Construction}
Given a rational $q = \frac{n}{m}$, we approximate its Adele embedding $\mathbf{A}(q)$ as follows:

\begin{enumerate}
    \item{\textbf{Real Component ($a_{\infty}$).}} 
    
    Represent $q$ on the real line as a length-$N$ vector:
    $$ a_{\infty} \mapsto [0, 0, \dots, q] \in \mathbb{R}^N $$

    \item{\textbf{$p$-adic Components ($a_p$).}}
    
    For each prime $p$ in a fixed set, compute the $N$-digit $p$-adic expansion of $q$:
    
    \begin{itemize}
        \item Solve the congruence $mx \equiv n \pmod{p^N}$.
        \item Use \textbf{Hensel's Lemma}~\cite{Hensel1908} to iteratively lift a solution modulo $p$ to one modulo $p^N$. This method is described in detail in Appendix~\ref{Hensel}
        \item Convert $x$ into base-$p$, yielding digits $[c_{N-1}, \dots, c_0] \in \mathbb{Z}_p^N$.
    \end{itemize}
    
    Thus, $$ a_p \mapsto [c_{N-1}, c_{N-2}, \dots, c_0]$$

    \item{\textbf{Final Adelic Vector.}} 
    
    Concatenate the real component and all chosen $p$-adic vectors:
    $$ \mathbf{A}(q) = \text{concat}(a_{\infty}, a_{p_1}, a_{p_2}, \dots, a_{p_{n_p}}) $$
    The resulting tensor has shape $(n_p+1)\times N$, where $n_p$ is the number of selected primes and $N$ is the digit precision.
\end{enumerate}

\section{Hensel's Lemma}
\label{Hensel}
In our method, we need to solve the congruence relation $mx \equiv n \pmod{p^N}$ to find the $p$-adic representation of a rational number $\frac{n}{m}$.
Hensel's Lemma provides a powerful and efficient way to achieve this.

At its core, Hensel's Lemma is an analogue of Newton's method for finding roots, but for modular arithmetic.
The core idea is "lifting": if you have an approximate solution to an equation modulo $p$, you can use that information to construct a more precise solution modulo $p^2$, then $p^3$, and so on, up to any desired power $p^N$.
This iterative refinement process is exactly what we need to determine the digits of a $p$-adic number to a specific precision.

More formally, one common version of Hensel's Lemma is stated for polynomial roots.

\begin{theorem}[Hensel's Lemma]
Let $f(x)$ be a polynomial with integer coefficients, and let $f'(x)$ be its derivative.
If an integer $r$ is a root of $f(x)$ modulo a prime $p$ (i.e., $f(r) \equiv 0 \pmod{p}$), and if the derivative at that point is non-zero modulo $p$ (i.e., $f'(r) \not\equiv 0 \pmod{p}$), then for any integer $k \ge 1$, there exists a unique integer $s$ modulo $p^k$ such that $f(s) \equiv 0 \pmod{p^k}$ and $s \equiv r \pmod{p}$.
\end{theorem}

\paragraph{Application to Our Method.}
In our case, we are not solving a general polynomial equation, but a linear congruence $mx \equiv n \pmod{p^N}$.
We can frame this as finding a root of the linear polynomial $f(x) = mx - n$.

\begin{itemize}
    \item The polynomial is $f(x) = mx - n$.
    \item Its derivative is $f'(x) = m$.
\end{itemize}

To apply Hensel's Lemma, we first need a solution $x_1$ modulo $p$. This requires solving $mx_1 - n \equiv 0 \pmod{p}$.
A solution exists if and only if $m \not\equiv 0 \pmod{p}$ (i.e., $p$ does not divide $m$).
The derivative condition, $f'(x_1) = m \not\equiv 0 \pmod{p}$, is the very same condition.

The iterative lifting process then works as follows: Suppose we have a solution $x_k$ such that $mx_k \equiv n \pmod{p^k}$.
We seek a new solution $x_{k+1}$ of the form $x_{k+1} = x_k + t \cdot p^k$ for some integer $t \in \{0, 1, \dots, p-1\}$.
We substitute this into the congruence for the next level:

\begin{align*}
    m(x_k + t\cdot p^k) &\equiv n \pmod{p^{k+1}} \\
    mx_k - n + m t\cdot p^k &\equiv 0 \pmod{p^{k+1}}
\end{align*}

Since $mx_k - n$ is a multiple of $p^k$, we can write it as $c \cdot p^k$ for some integer $c$. Substituting this gives:

\begin{align*}
    c \cdot p^k + m t \cdot p^k &\equiv 0 \pmod{p^{k+1}} \\
    c + m t &\equiv 0 \pmod{p} \quad (\text{after dividing by } p^k) \\
    t &\equiv -c \cdot m^{-1} \pmod{p}
\end{align*}

where $m^{-1}$ is the modular multiplicative inverse of $m$ modulo $p$, and $c = (mx_k - n)/p^k$.
This formula provides a direct way to compute the next "correction term" $t$, thereby lifting the solution from precision $k$ to $k+1$.
Our implementation programmatically applies this lifting from $k=1$ up to $N$.

\section{Dataset Details}
\label{app:dataset}

We conduct experiments on the \textbf{Algebraic Combinatorics Dataset (ACD) Repository}~\cite{chau2025machine}, a suite of nine research-level datasets designed to support the conjecturing phase of mathematical research in algebraic combinatorics. 
Each dataset pairs raw discrete combinatorial objects with mathematically meaningful labels, providing tasks that go beyond closed-form evaluation and emphasize pattern discovery for open mathematical problems.

\paragraph{Task Coverage.}
The ACD Repository spans diverse problems in algebraic combinatorics:
\begin{itemize}
    \item \textbf{Symmetric Group Characters}: Regression of irreducible symmetric group characters $\chi^\lambda_\mu$ for partitions $\lambda, \mu \vdash n$ ($n \in \{18,20,22\}$). This foundational result connects representation theory with combinatorial interpretations via Young diagrams.
    \newline\textbf{Example ($n=18$):} Input x is a pair of partitions. Output y is the integer character value.
    \begin{verbatim}
    x = {'Irreducible rep...': [5, 4, 3, ...], 
         'Conjugacy class': [7, 7, 4]}
    y = 0
    \end{verbatim}
    
    \item \textbf{mHeight Function}: Classification of the mHeight statistic on permutations ($n \in \{8,9,10\}$), a key tool in the recent proof of the Billey-Postnikov conjecture about Kazhdan-Lusztig polynomial coefficients.
    \newline\textbf{Example ($n=8$):} Input `x` is a permutation. Output `y` is its mHeight class.
    \begin{verbatim}
    x = {'Permutation': [2, 1, 5, 0, 7, 3, 6, 4]}
    y = 0 \end{verbatim}
    
    \item \textbf{Kazhdan-Lusztig Polynomial Coefficients}: Prediction of polynomial coefficients $P_{x,w}(q)$ indexed by permutation pairs ($n \in \{5,6,7\}$). These polynomials have deep connections to Schubert calculus and representation theory.
    \newline\textbf{Example ($n=5$):} Input `x` is a pair of permutations. Output `y` is a list of coefficients.
    \begin{verbatim}
    x = {'Permutation 1': [2, 1, 0, 3, 4], 
         'Permutation 2': [2, 1, 4, 3, 0]} 
    y = [1] \end{verbatim}
    
    \item \textbf{Robinson-Schensted-Knuth (RSK) Correspondence}: Recovery of permutations from pairs of standard Young tableaux ($n \in \{8,9\}$), testing whether models can learn this fundamental bijection in algebraic combinatorics.
    \newline\textbf{Example ($n=8$):} Input `x` is a pair of standard Young tableaux. Output `y` is the recovered permutation.
    \begin{verbatim}
    x = {'SYT 1': [[1, 3, 4, 6], [2, 8], ...], 
         'SYT 2': [[1, 2, 4, 6], [3, 5], ...]} 
    y = [2, 7, 3, 8, 5, 6, 4, 1] \end{verbatim}
    
    \item \textbf{Schubert Polynomial Structure Constants}: Prediction of structure constants $c^\gamma_{\alpha,\beta}$ in the expansion $S_\alpha S_\beta = \sum_\gamma c^\gamma_{\alpha,\beta} S_\gamma$ ($n \in \{4,5,6\}$). Finding combinatorial interpretations of these constants is a major open problem.
    \newline\textbf{Example ($n=4$):} Input `x` is a triplet of indices. Output `y` is the integer structure constant.
    \begin{verbatim}
    x = {'Lower index 1': [3, 1, 4, 2], 
         'Lower index 2': [1, 4, 3, 2],
         'Upper index': [3, 7, 4, ...]} 
    y = 0 \end{verbatim}
    
    \item \textbf{Grassmannian Cluster Algebras}: Binary classification of whether semistandard Young tableaux of shape $3 \times 4$ with entries from $\{1,\ldots,12\}$ index cluster variables in $\text{Gr}(3,12)$.
    \newline\textbf{Example:} Input `x` is a semistandard Young tableau. Output `y` is a binary label.
    \begin{verbatim}
    x = {'SSYT': [[1, 1, 5, 7], [4, 5, 7, 9], ...]} 
    y = 0 \end{verbatim}
    
    \item \textbf{Lattice Path Partial Orders}: Classification of covering pairs in lattice paths from $(0,0)$ to $(n,n-1)$ under Lagrange vs. matching orderings ($n \in \{10,11,12,13\}$), motivated by connections to number theory.
    \newline\textbf{Example ($n=10$):} Input `x` is a pair of paths. Output `y` is an integer label.
    \begin{verbatim}
    x = {'Lattice path 1': [1, 1, 0, ..., 0, 1, 0], 
         'Lattice path 2': [1, 1, 0, ..., 1, 0, 0]} 
    y = 0 \end{verbatim}
    
    \item \textbf{Quiver Mutation Classes}: Classification of 11-vertex quivers into seven mutation equivalence classes (A$_{11}$, D$_{11}$, E$_{11}$, BD$_{11}$, BE$_{11}$, DE$_{11}$, BB$_{11}$), where determining mutation equivalence is an open algorithmic problem.
    \newline\textbf{Example:} Input `x` is a flattened adjacency matrix. Output `y` is the mutation class name.
    \begin{verbatim}
    x = {'Adjacency matrix': [0, 0, 0, ..., 1, 0, ...]} 
    y = 'A_11' \end{verbatim}
    
    \item \textbf{Weaving Patterns}: Binary classification of $n \times (n-1)$ matrices with entries in $\{1,\ldots,n\}$ as valid weaving patterns ($n \in \{6,7\}$), related to reduced decompositions of the longest permutation.
    \newline\textbf{Example ($n=6$):} Input `x` is a flattened matrix. Output `y` is a binary label.
    \begin{verbatim}
    x = {'Matrix': [5, 4, 3, 2, 1, 6, 2, 3, ...]} 
    y = 1 \end{verbatim}
\end{itemize}

\paragraph{Scale and Characteristics.}
The datasets exhibit substantial scale variation, ranging from thousands to over 10 million examples, with dataset sizes typically growing exponentially in the parameter $n$. 
Class imbalance is prevalent across tasks—for instance, zero-valued structure constants vastly outnumber non-zero ones, and most quiver mutation classes are represented unequally. 
The distributions often exhibit heavy tails (e.g., symmetric group characters are concentrated around zero with long tails), making standard accuracy metrics potentially misleading.

\paragraph{Mathematical Significance.}
Unlike typical machine learning benchmarks, these datasets represent either:
\begin{enumerate}
    \item \textbf{Foundational results} where efficient algorithms exist (e.g., RSK correspondence, symmetric group characters) but learning them from examples tests a model's ability to discover mathematical structure;
    \item \textbf{Open problems} where no complete characterization is known (e.g., Schubert polynomial structure constants, quiver mutation equivalence for general types), making them genuine research-level challenges.
\end{enumerate}

\paragraph{Tasks Used in Our Study.}
From this collection, we focus on six classification tasks that span the breadth of combinatorial objects and mathematical contexts: mHeight function, Grassmannian cluster algebras, lattice path partial orders, quiver mutation classes, weaving patterns, and Schubert polynomial structure constants. 
These tasks were selected to provide diverse input representations (permutations, tableaux, matrices, graphs) and varying degrees of class imbalance.
Full dataset statistics, generation procedures, and train/test splits are provided with the ACD Repository.

\section{Training Details}
\label{app:training}
\paragraph{Optimizer and Objective.}
We use the Adam optimizer with default $\beta$ values $(0.9, 0.999)$ and weight decay set to $0$. 
The training objective is Cross-Entropy Loss for all classification tasks.

\paragraph{Learning Rate Scheduler.}
We employ a cosine annealing learning rate scheduler \verb|CosineAnnealingLR|. 
For each task, the maximum number of epochs is set as $T_{\max}$, ensuring that the learning rate decays smoothly over the full training run. 
Initial learning rates are task-specific, as summarized in Table~\ref{tab:training_hparams}.

\paragraph{Batching and Sampling.}
All experiments use mini-batches of size 2048.  
To address severe class imbalance, we adopt a \verb|WeightedRandomSampler|, where the sampling weight for class $i$ is proportional to $\frac{1}{\sqrt{n_i}}$, with $n_i$ denoting the number of samples in class $i$. 
This strategy ensures minority classes are not neglected while avoiding over-amplification of extremely rare labels.

\paragraph{Training Epochs.}
Each task is trained with its own maximum epoch count (see Table~\ref{tab:training_hparams}). 
Validation is performed at the end of each epoch, and the checkpoint with the best validation accuracy is selected for reporting final results.

\paragraph{Hardware.}
All models are trained on a single NVIDIA RTX 5090 GPU (32GB). 
Experiments are implemented in PyTorch (version \texttt{2.6.0a0+df5bbc09d1.nv24.11}). 

\paragraph{Hyperparameter Summary.}
Table~\ref{tab:training_hparams} lists the main hyperparameters and per-task settings used in our experiments.

\begin{table}[t]
\centering
\begin{tabular}{llcc}
\toprule
\textbf{Task} & \textbf{$n$} & \textbf{Learning Rate} & \textbf{Epochs} \\
\midrule
\multirow[t]{4}{*}{\raggedright Lattice paths} 
    & $n=10$        & $7.7\mathrm{e}{-5}$ & 100 \\
    & $n=11$        & $1\mathrm{e}{-3}$ & 100 \\
    & $n=12$        & $1\mathrm{e}{-3}$ & 20 \\

\midrule
\multirow[t]{2}{*}{\raggedright Weaving patterns} 
    & $n=6$         & $2\mathrm{e}{-5}$ & 100 \\
    & $n=7$         & $1\mathrm{e}{-4}$ & 100 \\

\midrule
\multirow[t]{2}{*}{\raggedright Quiver mutation classes} 
    &               & $8.8\mathrm{e}{-5}$ & 100 \\

\midrule
\multirow[t]{3}{*}{\raggedright Grassmannian cluster algebras} 
    &               & $6\mathrm{e}{-5}$ & 100 \\

\midrule
\multirow[t]{3}{*}{\raggedright Schubert constants} 
    & $n=4$         & $1.7\mathrm{e}{-5}$ & 200 \\
    & $n=5$         & $8\mathrm{e}{-5}$ & 100 \\
    & $n=6$         & $5\mathrm{e}{-5}$ & 50 \\

\midrule
\multirow[t]{3}{*}{\raggedright mHeight} 
    & $n=8$         & $3\mathrm{e}{-4}$ & 100 \\
    & $n=9$         & $6\mathrm{e}{-4}$ & 100 \\
    & $n=10$        & $7.3\mathrm{e}{-5}$ & 30 \\
\bottomrule
\end{tabular}
\caption{Per-task training hyperparameters. CosineAnnealingLR scheduler uses $T_{\max} =$ maximum epoch for each configuration. Batch size is fixed to 2048 for all tasks. Learning rates were selected via Optuna-based hyperparameter search.}
\label{tab:training_hparams}
\end{table}

\section{Test Loss Table}
\cref{tab:main_loss} lists the test losses per-task corresponding to the accuracy results reported in \cref{results}. 
While accuracy provides a direct measure of task performance, the loss values offer an additional perspective on model calibration and confidence. 
Consistent with the accuracy gains, AOE achieves lower or comparable losses across nearly all tasks, reinforcing that its improvements are not limited to classification correctness but extend to the stability of the underlying optimization.

\begin{table}[ht]
\centering
\caption{Final test loss comparison on the Algebraic Combinatorics benchmark. Lower is better.}
\label{tab:main_loss}
\begin{tabular}{llrr}
\toprule
\multicolumn{2}{c}{\textbf{Dataset Task}} & \multicolumn{2}{c}{\textbf{Models}} \\
\cmidrule(lr){1-2} \cmidrule(lr){3-4}
\textbf{Task} & \textbf{$n$} & \textbf{Baseline} & \textbf{Ours} \\
\midrule

\multirow[t]{3}{*}{Lattice paths} 
    & 10 & 0.7045 & 0.5957 \\
    & 11 & 0.6893 & 0.6454 \\
    & 12 & 0.6845 & 0.4658 \\
\midrule
\multirow[t]{2}{*}{Weaving patterns} 
    & 6  & 0.6910 & 0.6897 \\
    & 7  & 0.6933 & 0.0001 \\
\midrule
Quiver mutation classes &   & 0.9873 & 0.2247 \rule{0pt}{2.5ex}\\
\midrule
Grassmannian cluster algebras &   & 0.2422 & 0.0827 \rule{0pt}{2.5ex}\\
\midrule
\multirow[t]{3}{*}{Schubert polynomials} 
    & 4  & 0.6937 & 0.3052 \\
    & 5  & 1.0910 & 0.1093 \\
    & 6  & 0.8031  & 0.0020 \\
\midrule
\multirow[t]{3}{*}{mHeight} 
    & 8  & 0.4740 & 0.1190 \\
    & 9  & 0.4329 & 0.0230 \\
    & 10 & 0.3998 & 0.0105 \\

\bottomrule
\end{tabular}
\end{table}


\end{document}